\documentclass[review]{elsarticle}

\usepackage{graphicx}
\usepackage{amsmath,amssymb,mathtools}
\usepackage{graphicx,subfigure}
\usepackage{amsmath}
\usepackage{bm}
\DeclareMathOperator{\vb}{\mathbf{b}}
\DeclareMathOperator{\vc}{\mathbf{c}}
\DeclareMathOperator{\ve}{\mathbf{e}}
\DeclareMathOperator{\vv}{\mathbf{v}}

\DeclareMathOperator{\vu}{\mathbf{u}}
\DeclareMathOperator{\vt}{\mathbf{t}}
\DeclareMathOperator{\mT}{\mathbf{T}}
\DeclareMathOperator{\valpha}{\mathbf{\alpha}}
\DeclareMathOperator{\vone}{\mathbf{1}}
\DeclareMathOperator{\E}{\mathbb{E}}
\DeclareMathOperator{\R}{\mathbb{R}}
\DeclareMathOperator{\Eps}{\mathcal{E}}
\DeclareMathOperator{\Tau}{\mathcal{T}}
\DeclareMathOperator{\scC}{\mathcal{C}}
\DeclareMathOperator{\scU}{\mathcal{U}}

\usepackage{lineno, hyperref}
\usepackage{booktabs}
\usepackage{tabstackengine}

\nolinenumbers

\usepackage{xcolor}

\journal{Journal of \LaTeX\ Templates}

\begin{document}

\begin{frontmatter}

\title{Information Avoidance and Overvaluation in Sequential Decision Making under Epistemic Constraints
}


\author{Shuo Li \&, Matteo Pozzi}

\author{~~\textit{Carnegie Mellon University}
\newline - \newline manuscript submitted to Reliability Engineering and System Safety}



\begin{abstract}
Decision makers involved in the management of civil assets and systems usually take actions under constraints imposed by societal regulations.
Some of these constraints are related to epistemic quantities, as the probability of failure events  and the corresponding risks.
Sensors and inspectors can provide useful information supporting the control process (e.g. the maintenance process of an asset), and decisions about collecting this information should rely on an analysis of its cost and value.
When societal regulations encode an economic perspective that is not aligned with that of the decision makers, the Value of Information (VoI) can be negative (i.e., information sometimes hurts), and almost irrelevant information can even have a significant value (either positive or negative), for agents acting under these epistemic constraints.
We refer to these phenomena as Information Avoidance (IA) and Information OverValuation (IOV).

In this paper, we illustrate how to assess VoI in sequential decision making under epistemic constraints (as those imposed by societal regulations), by modeling a Partially Observable Markov Decision Processes (POMDP) and evaluating non optimal policies via Finite State Controllers (FSCs).
We focus on the value of collecting information at current time, and on that of collecting sequential information, we illustrate how these values are related and we discuss how IA and IOV can occur in those settings. 
\end{abstract}


\end{frontmatter}


\section{Introduction}

We investigate how to assess the Value of Information (VoI) under epistemic constraints in sequential decision making, we discuss how the VoI can be increased or decreased by these constraints, even becoming negative.
We refer to the VoI being negative as a case of Information Avoidance (IA) and to the VoI being largely increased by the constraints as a case of Information OverValuation (IOV).
Our target application is the management process of urban and civil assets and systems, including operation and maintenance.

Asset management is a sequential decision making process, where a decision maker (that we call an ``agent") takes periodic actions under uncertainty, with a goal that can be modeled as the minimization of a metric, e.g. the expected long-term costs~\cite{frangopol2004probabilistic}.
When the current state is not perfectly observable, this process can be formulated a Partially Observable Markov Decision Process (POMDP)~\citep{papakonstantinou2014planning, memarzadeh2015optimal}.
POMDP and MDP were extensively applied to operation and maintenance, e.g. of bridges~\cite{scherer1994markovian, jiang2000optimal}, of electric-power systems~\cite{platis1996performability, dhople2013set, compare2020partially}, of road pavements~\cite{smilowitz2000optimal, robelin2007history, gao2013markov}, 
of structural systems~\cite{papakonstantinou2014planning2}, 
of coastal protection~\cite{PozziMemarzadehKlima2017}, of wind mills~\cite{byon2010season, memarzadeh2014optimal, nielsen2015risk} and large infrastructure systems ~\cite{andriotis2019managing}.

Sensors and inspecting technologies can reduce the uncertainty of the control process, and the overall cost.
However, decisions about collection of information should be taken based on its cost and value.
The VoI is a utility-based metric introduced by the seminal work of~\cite{raiffa1961applied}, and it measures the impact of information in expected utility and loss.
VoI analysis has been applied to inspection scheduling by \citep{STRAUB2005}, integration of structural monitoring by \citep{pozzi2011assessing, qin2015value,  thons2018value, straub2014value, zonta2014value} and multi-step maintenance by \citep{goulet2015pre}, sensor placement \cite{malings2016value, hoseyni2021optimal} and reviewed by \cite{zhang2021value}.
The VoI of a monitoring system, in sequential decision making, can be derived as the difference between the expected discounted long-term cost without and with the information. 
Analysis of VoI in sequential decision making using POMDPs is illustrated by \cite{memarzadeh2016value}.
It can refer to the value of current information (collectable at current time), or to that of a flow of information (collectable at present and future times).
The value of current information depends on the availability of future information, and \cite{memarzadeh2016integrated} introduces some limit-cases assumptions, that we will also adopt in our work.
Other more complicated assumptions are proposed by \cite{memarzadeh2016value}.
The value of flow of information is explored by \cite{srinivasan2013value, memarzadeh2016value, li2019makes} and \cite{andriotis2019value} (which also investigates the value of current information, and proves properties and inequalities related to optimal policies and minimum costs).

Without external constraints, rational agents always prefer to collect free information, according to principle that ``information never hurts"~\cite{cover2006elements}, and almost irrelevant information has small VoI.
However, asset owners and managers of infrastructure components usually act under external constraints, as those imposed by society via regulations, to guarantee an adequate safety level.
We qualify these constraints as ``epistemic", because they refer to an epistemic quantities, as the probability of failure. 
These constraints affect not only the management actions, but also the attitude towards information.
For example, decision about installing sensors and inspecting assets can be affected by these constraints: it may be the case that the installment can be convenient without regulation, but it is not convenient anymore under the regulation.

Paper \citep{pozzi2020information} shows the basic properties of VoI do not hold for agents acting under epistemic constraints.
As in the example reported in that paper, consider the case of a regulation requiring to perform a costly maintenance action when the asset's failure probability is above an admissible threshold, an agent willing to take risks much higher than those accepted by society, and a current failure probability just below the threshold.
In this case, the agent may prefer to avoid collecting free information about the asset's state (and even to pay to avoid it), as she is worried that the information can increase the posterior failure probability above the threshold, forcing her to take the expensive repair she considers unnecessary.
IA (and IOV) has been investigated, in the context of asset management, by
\cite{pozzi2017negative, pozzi2020information, bolognani2019quantifying, verzobio2021quantifying}, in that of non-cooperative games by \cite{bertschinger2013information}, in that of social science by \cite{sweeny2010information, golman2017information}.

In this article, we extend that analysis of the one-shot problem to sequential decision processes.
As the policy imposed via regulation by society to the agent is sub-optimal for the latter, a key task in our analysis is to evaluate sub-optimal policies in sequential decision making, and we adopt methods based on Finite State Controllers (FSCs), as illustrated in \cite{li2021predicting}.

Following \cite{pozzi2020information}, we justify the  societal constraints showing how the policy imposed to the agent is optimal under the costs assessed from the societal standpoint.
When assessing the VoI, we mainly focus on the agent's costs, assuming that ``exploratory" actions, e.g. about installing sensors or inspecting components, are taken by the agent without any societal constraint.
Our findings extends those of \cite{pozzi2020information} and, following the approach of that paper, they can also be related to the design of a mechanism to alleviate IA and IOV.
However, in this paper, we do not explicitly adapt the design of mechanisms to sequential decision making.

The rest of the article is organized as follows.
Sec.~\ref{ProblStat} presents the problem and introduces the formulation of POMDPs.
Sec.~\ref{PolEval} discusses how to evaluate and optimize policies, Sec.~\ref{VoI} how to define and assess the VoI in sequential decision making and Sec.~\ref{OptVal} refers to the special case of optimal decision making.
Sec.~\ref{ModSocReg} discusses how societal regulations are defined and Sec.~\ref{IAOE} illustrates how IA and IOV are related to the geometric properties of the value function quantifying the long-term expected cost of the process, which is affected by the epistemic constraints.
Sec.~\ref{Examples} presents examples and a parametric analysis to investigate how IA and IOV depend on features of the process, e.g. the information precision and the degradation rate, and Sec.~\ref{Concl} draws conclusions.
The appendix provides a proof of a statement about the relation between value of current and of flow of information, and details on point-based value iteration algorithms for POMDPs.

\section{Problem formulation}~\label{ProblStat}
\subsection{Engineering problem statement}
The main question of this article is: ``how can we model the attitude toward information for agents taking sequential decisions under epistemic constrains?"
A more specific related question is ``when is the VoI negative, in sequential decision making?"
To answer these questions, we consider an agent taking sequential decisions about the management of a system. During the process, the agent pays costs for maintaining the system, and penalties for the system  malfunctioning. 
While the agent's goal is to minimize the long-term expected accumulated cost, she has to follow societal regulations, e.g. she must ``repair" when the probability of the system being damaged is high.
Note that we assume that all information are public, so the agent must share any collected information with the regulator. More generally, the agent and the regulator share the same ``belief" about the system state, and the same evolution model.

Background information are available to the agent, and the policy is defined on the inferred system condition, accounting for this information.
Now, additional information is also available, and the agent can take or avoid such source.
To decide about the collection of this additional information, the agent must assess its value.
If such VoI is negative, information will be avoided, even if free of cost.
We focus on VoI for additional information available only at current time, and for information available also in the future.

\subsection{The stochastic decision process}
\begin{figure}
    \centering  \includegraphics[scale=0.52]{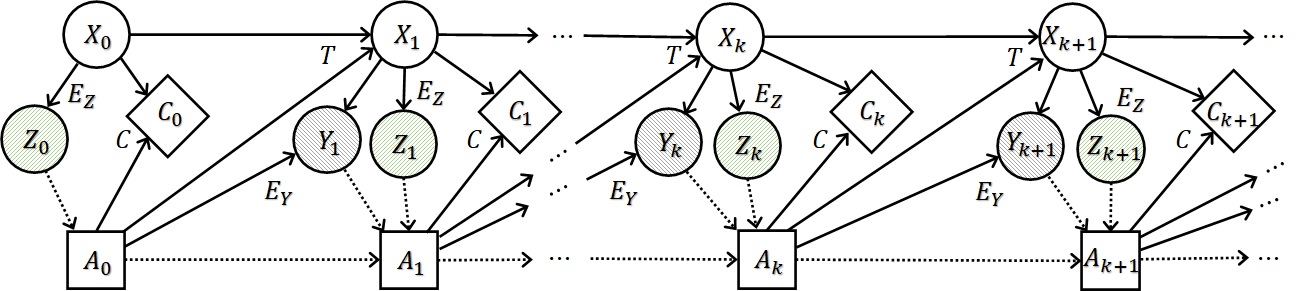}
    \caption{Influence diagram for a POMDP, with background and additional information.}
    \label{inGraph}
\end{figure}

We model the evolution of a controlled system as a POMDP, as illustrated in Fig.~\ref{inGraph}. 
Time is discretized in a sequence of evenly spaced instants $\{t_0,t_1,\dots\}$, when control actions are taken, costs paid and observations collected, and variables $x_k, a_k, c_k$ represent the physical state of the system, the taken action and cost paid at time $t_k$.
The physical state belongs to finite domain $\Omega_X = \{1,2,\dots, n_X\}$ and the agent selects actions in domain $\Omega_A = \{1,2,\dots, n_A\}$ (hence, $n_X$ and $n_A$ are the cardinality of the state and of the action domain, respectively).

The state evolution follows a Markov process, governed by transition function $T(x,a,x') = \mathbb{P}[x_{k+1}=x' | x_k = x, a_k = a]: [\Omega_X \times \Omega_A \times \Omega_X]\rightarrow [0,1]$.
The immediate cost is defined by cost matrix $C(x,a): [\Omega_X \times \Omega_A]\rightarrow \R$, depending on current time and action, so that cost at time $t_k$ is $c_k = C(x_k, a_k)$.

Before an action is taken, some (possibly noisy) measures of the state are available to the agent.
At time $t_k$, ``background" observation $y_k$ belongs to domain $\Omega_Y = \{1,2,\dots,n_Y\}$ (where $n_Y$ is the cardinality of $\Omega_Y$).
The relation between state and background observation is defined by emission function $E_Y(x,a,y) = \mathbb{P}[y_k=y|x_k=x, a_{k-1}=a]: [\Omega_X\times\Omega_A\times\Omega_Y]\rightarrow[0,1]$.
Furthermore, at any time $t_k$, additional information $z_k$ is available, and we aim at assessing its value.
We assume that the additional information is emitted by function $E_Z(x, z) =  \mathbb{P}[z_k=z|x_k=x]: [\Omega_X\times\Omega_Z] \rightarrow [0,1]$, on domain $\Omega_Z = \{ 1,2, \dots, n_Z \}$ (where $n_Z$ is the cardinality of $\Omega_Z$), independently on actions.
Background and additional information are combined in joint information $w=\{y, z\}\in\Omega_W$ and, if observations $y$ and $z$ are independent conditional to hidden state $x$, then the emission probability of $w$ is $E_W(x,a,w) =  \mathbb{P}[w_k=w|x_k=x,a_{k-1}=a]: [\Omega_X\times\Omega_A\times\Omega_W] \rightarrow [0,1]$, is:
\begin{equation}
    \label{eq:E_W}
    E_W(x,a,w) = E_Y(x,a,y) E_Z(x,z)
\end{equation}
where the cardinality of the domain $\Omega_W = \{ 1,2, \dots, n_W\}$ of the joint information is $n_W = n_Y \cdot n_Z$.

Knowledge about the current state is represented by a probability distribution (called ``belief"), listed a $n_X$-dimensional stochastic column vector $\vb = [b(1), b(2), \dots, b(n_X)]^\top\in\Omega_B$, where convex set $\Omega_B=[0,1]^{n_X}: \vone^\top\vb=1$  is the domain of  beliefs (and $\vone$ is a column vector of ones).

\subsection{Belief updating}
~\label{Belief_updating}
If additional information is collected at current time, then the current belief is updated, following Bayes' formula.
To process value $z$, we define emission vector $\ve_Z(z) = [ E_Z(1, z), E_Z(2,z), \dots, E_Z(n_X, z)]^\top$ as column $z$ of the emission table $E_Z$.
The probability of observing $z$ from belief $\vb$ is $\Eps_Z(\vb,z) = \ve_Z(z)^\top \vb: [\Omega_B\times\Omega_Z] \rightarrow [0, 1]$.
The posterior belief $\vu_Z$ is computed using Bayes' formula, as 
\begin{equation}
    \label{eq:u_Z}
    \mathbf{u}_Z(\vb,z) = \frac{ \text{Diag}[\ve_Z(z)] ~\vb }{\Eps_Z(\vb,z)} ~~: [\Omega_B\times\Omega_Z] \rightarrow\Omega_B
\end{equation}
where $\text{Diag}[\vv]$ is a square matrix with vector $\vv$ on the diagonal, and all zero entries out of the diagonal.

Similarly, we can define operators $\ve_Y$, $\vu_Y$ $\Eps_Y$, using emission matrix $E_Y$ for background information, and $\ve_W$, $\vu_W$,  $\Eps_W$, using emission matrix $E_W$ for the joint information (these operators also depend on the action selected at previous step).

The one-step ahead transition operator, $\vt(\vb,a):[\Omega_B \times\Omega_A] \rightarrow \Omega_B$ computes, using total probability rule, the belief for the state at next step, without processing any new information.
Entry $i$ of this operator is:
\begin{equation}
    \label{eq:t_b}
    \vt(\vb,a)(i) = \sum_{j=1}^{n_X} T(j,a,i) b(j)
\end{equation}
and the operator can be expressed as: $\vt(\vb,a)=\mT_a^\top\vb$, where entry $\{i,j\}$ in square stochastic transition matrix $\mT_a$ is $T(i,a,j)$.
Transition and updating operators can be integrated into a single operator that, as a function of current belief $\vb$, current action $a$ and information at next step, compute the updated belief for next state.
When only background information $y$ is available at next step, the corresponding operator, $\Tau_Y$, is:
\begin{equation}
    \label{eq:T_Y}
    \Tau_Y(\vb,a,y) = \vu_Y\big(\vt(\vb,a),y\big) ~:~ [\Omega_B\times\Omega_A\times\Omega_Y] \rightarrow\Omega_B
\end{equation}
Similarly, we can define operator $\Tau_W$ for transition and updating when also the additional information is available, using updating operator $\vu_W$ instead of $\vu_Y$ in Eq.\eqref{eq:T_Y}.

Hence, initial belief $\vb_0$, at time $t_0$, evolves to belief $\vb'_0=\vu_Z(\vb_0,z_0)$ if additional information $z_0$ is collected and processed, while it does not change is no additional information is collected (i.e., $\vb'_0=\vb_0$).
Then, action $a_0$ is selected according to a policy applied to belief $\vb'_0$, and belief evolves to $\vb_1=\Tau_Y(\vb'_0,a_0,y_1)$ if only background information $y_1$ is available at next step, and to $\vb_1=\Tau_W(\vb'_0,a_0,w_1)$ if also additional information $z_1$ is collected.
Then, at any time $t_k$, current belief belief $\vb_k$ is updated to belief at next time $\Tau_W(\vb_k,a_k,w_{k+1})$ if additional information $z_{k+1}$ is collected and processed, and to $\Tau_Y(\vb_k,a_k,y_{k+1})$ if it is not.

\section{Policy evaluation and pre-posterior analysis}~\label{PolEval}
In POMDPs, the belief is a ``sufficient statistic" of the process, and hence actions should be selected as a function of the belief.
If the process is stationary and related to an infinite horizon, then the optimal policy is also stationary.
A stationary policy is a function, mapping current belief into action:
$\pi(\vb) : \Omega_B \rightarrow \Omega_A$.
We are interested in evaluating a policy, i.e. in assessing the expected discounted cumulative cost (that here we call the ``value") following that policy forever.
Value $V$ can be represented as a function in the belief domain, which depends on the adopted policy $\pi$, defined as:
\begin{equation}
    \label{eq:V_pi}
    V^\pi(\vb) = \E \Big[\sum_{k=0}^{\infty} \gamma^k c_k | \vb_0 = \vb\Big]~:~ \Omega_B \rightarrow \R
\end{equation}
where $\gamma$ is the one-step discount factor, and the sequence of costs is for trajectories when actions are selected following the policy $\pi$.

\subsection{Value functions}
To compute value functions, we redefine transition operators as related to specific policy $\pi$.
The transition to the next step becomes:
\begin{equation}
    \label{eq:t_pi}
    \vt^\pi(\vb)=\vt\big(\vb,\pi(\vb)\big) ~:~ \Omega_B\rightarrow\Omega_B
\end{equation}
and that also including the updating process due to background information is:
\begin{equation}
    \label{eq:T_Y_pi}
    \Tau_Y^\pi(\vb,y)=\Tau_Y(\vb,\pi(\vb),y) ~:~ [\Omega_B\times\Omega_Y] \rightarrow\Omega_B
\end{equation}
Similarly, we define operator $\Tau_W^\pi$, using operator $\Tau_W$ instead of $\Tau_Y$, to include also the processing of additional information.

If $\vb$ represent the current belief, after any current information has been processed, and if only background information will be available in the future, then value function $V^\pi_Y$ of Eq.\eqref{eq:V_pi}, following policy $\pi$, can be expressed by a recursive Bellman equation ~\cite{Bertsekas2005} as:
\begin{equation}
    \label{eq:V_pi_Y}
    V_Y^\pi(\vb) = \scC^\pi(\vb) + \gamma\sum_{y=1}^{n_Y} \Eps_Y\big(\vt^\pi(\vb),y\big)~ V_Y^\pi\big(\Tau_Y^\pi(\vb,y)\big)
\end{equation}
where $\scC^\pi(\vb) = \sum_{i=1}^{n_X} C\big(i,\pi(\vb)\big)b(i)$ is the immediate expected cost following policy $\pi$ (clearly, the operator is linear in belief $\vb$).
We use subscript $Y$ to remind that only background information is available.

Similarly, we define value function $V_W^\pi$, assuming that additional information is always available in the future.
To do so, we use operators $\Eps_W$ and $\Tau_W^\pi$ in Eq.\eqref{eq:V_pi_Y} (instead of $\Eps_Y$ and $\Tau_Y^\pi$), and the sum is on variable $w$, ranging from $1$ to $n_W$.

\subsection{Numerical approaches for policy evaluation}
\label{NumApprPolEv}
To compute the value function, we need to adopt a numerical scheme to implement Eq.\eqref{eq:V_pi_Y}. We follow \cite{li2021predicting} to evaluate policies which are non optimal, using Finite state controllers (FSCs).
A FSC follows a conditional plan, to take actions processing a sequence of the collected information.
When only background information is available, the current belief is summarized by an ``inner state", in finite discrete domain $\Omega_H = \{1, 2, \dots, n_H\}$, by function $m_Y(\vb):\Omega_B\rightarrow\Omega_H$.
At time $t_k$, current inner state $h=h_k$ is updated into next inner state $h'=h_{k+1}$ depending on the background $y$ information at next step, via updating function: $h' = \eta^\pi_Y(h, y): \Omega_H \times\Omega_Y\rightarrow \Omega_H$.
Current action depends on current inner state, via function $\pi(h):\Omega_H\rightarrow\Omega_A$ (we use the same letter for indicating the policy as a function of the belief and as a function of the inner state, as the meaning will be clear from the context).
The Grid Method (GM) and the Optimal Region Method (ORM), as summarized by ~\cite{li2021predicting}, are alternative approaches for implementing a FSC in POMDP.
Details on how to identify functions $\eta^\pi_Y$ and $\pi$, using point-based value iterations methods, are reported in ~\ref{App_PBVI}.

Following~\cite{li2021predicting}, we introduce a joint state $s = \{x, h\}$, merging ``physical and inner state, defined in joint domain $\Omega_S = \Omega_X\times\Omega_H$, of cardinality $n_S=n_X\cdot n_H$.
The evolution of the joint state is Markovian, and the transition probability is: 
\begin{equation}
    \label{eq:T_tilde_pi_Y}
    \tilde{T}^\pi_Y(s,s') = T(x,\pi(h),x') \sum_{y=1}^{n_Y} E_Y(x',\pi(h),y)~\mathbb{I}\big[h'= \eta_Y^\pi(h, y)\big]
\end{equation}
where $s' = \{x', h'\}$ is the joint state at next step, and $\mathbb{I}[\cdot]$ is the indicator function. These transition probabilities can be listed in matrix $\tilde{\mathbf{T}}^\pi_Y$, of size $(n_S\times n_S)$.
Immediate cost, as a function of current joint state $s$, is $\tilde{c}^\pi(s)=C\big(x,\pi(h)\big)$,
and it can be listed in column vector $\tilde{\vc}^\pi$, of size $n_H$.
The value, as a function of current joint state $s$, is $\tilde{v}^\pi(s)$, and it can also be listed in column vector $\tilde{\vv}^\pi$, of size $n_S$.
Such value can be expressed recursively, as in Eq.\eqref{eq:V_pi_Y}, as:
\begin{equation}
    \label{eq:PolEval}
    \tilde{\vv}^\pi_Y = \tilde{\vc}^\pi + \gamma~\tilde{\mathbf{T}}^\pi_Y~ \tilde{\vv}^\pi_Y
\end{equation}
This is a system of $n_S$ linear equation in $n_S$ unknowns, that can be solved using traditional methods for linear algebra (when $\gamma<1$ the solution is unique).

When vector $\tilde{\vv}^\pi_Y$ is computed, function $V^\pi_Y$ is computed using map $m_Y$, from belief to inner state.
We identify a $n_X$-dimensional column vector $\tilde{\vv}^\pi_{Y,(h)}$, extracting from vector $\tilde{\vv}^\pi_Y$ the entries related to inner state $h=m_Y(\vb)$. This is related to the concept of $\alpha$-vector, as discussed in Sec.~\ref{OptVal} and in ~\ref{App_PBVI}.
Then we express the value as a linear function of the belief:
\begin{equation}
    \label{eq:pieceWiseLinValue}
    V^\pi_Y(\vb) = \sum_{i=1}^{n_X} \tilde{v}^\pi_{Y,(h)}(i)~b(i) = {\vv^\pi_{Y,m_Y(\vb)}}^\top \vb
\end{equation}
Hence, function $V^\pi_Y$ is linear in the subset of belief domain $\Omega_B$ where map $m_Y$ is constant, i.e. in the subset of beliefs mapped into the same inner state.
Instead, at the border between subsets, where map $m_Y$ is discontinuous, function $V^\pi_Y$ can also be discontinuous as noted, in a similar context, in \cite{li2021predicting}.

Similarly, we can evaluate a policy when additional information is always available in the future.
To do so, we use emission matrix $E_W$ and updating function $\eta_W^\pi$ in Eq.\eqref{eq:T_tilde_pi_Y},  and get matrix $\mathbf{\tilde{T}}^\pi_W$ and then, recursively, vector $\mathbf{\tilde{v}}^\pi_W$ using the approach of Eq.\eqref{eq:PolEval}.

\subsection{Pre-posterior value functions}
Given a value function,
the corresponding pre-posterior function includes the availability of current additional information.
If no future additional information will be available, the value function is $V^\pi_Y$, and the pre-posterior function is:
\begin{equation}
    \label{eq:PrePostValue}
    \scU^\pi_Y(\vb) = \E_{Z_0} \big[V^\pi_Y\big( \vu_Z(\vb,z_0)\big)\big] = \sum_{z=1}^{n_{Z}} \Eps_Z(\vb,z)~ V^\pi_Y\big(\vu_Z(\vb,z)\big)
\end{equation}
Similarly, when additional information is always available in the future, we define pre-posterior function $\scU^\pi_W$, using value function $V^\pi_W$ instead of $V^\pi_Y$ in Eq.\eqref{eq:PrePostValue}.
Note that the number of inner states can be different with and without additional information.

\section{Value of information}
~\label{VoI}
\subsection{Selection of policies}
To assess the impact of additional information, we need to define what policy the agent follows.
We consider two policies: the agent adopts policy $\pi_A$ when only background information will be available in the future, and policy $\pi_B$ if additional information will always be available.

We are interested in the value of information collected at the current time and of a that of sequence of information collected from the current step and forever in the future.
Following \cite{memarzadeh2016integrated}, we start assessing the Value of Current Information, that is the change in the value function, because observation $z_k$ is collected at current time $t_k$.
As noted in that paper, this change depends on the assumption on the availability of future information. 
In the ``pessimistic" assumption, no additional information will ever be available in future steps.
If so, the pessimistic value of current information, $VoI_\text{C}^\text{P}$, is:
\begin{equation}
    \label{eq:VoI_C}
    VoI_\text{C}^\text{P}(\vb) = V^{\pi_A}_Y(\vb) - \mathcal{U}^{\pi_A}_Y(\vb)
\end{equation} 
The ``optimistic" assumption claims that additional information will be available in all future steps (regardless the acquisition of current information), and the corresponding value is:
\begin{equation}
    \label{eq:VoI_O}
    VoI_\text{C}^\text{O}(\vb) = V^{\pi_B}_W(\vb) - \scU^{\pi_B}_W(\vb)
\end{equation}
While other intermediate assumptions has also been proposed and investigated \cite{memarzadeh2016value}, this paper focuses only on these two limit assumptions.

Again following \cite{memarzadeh2016value}, we define the Value of Flow of Information, $VoI_\text{F}$, as that of obtaining additional information for all steps, including the current one: 
\begin{equation}
    \label{eq:VoI_F}
    VoI_\text{F}(\vb) = V^{\pi_A}_Y(\vb) - \scU^{\pi_B}_W(\vb) 
\end{equation}
We prove in \ref{App_Proof} that this value can be expressed by a recursive Bellman equation, similar to Eq.\eqref{eq:V_pi_Y}:
\begin{equation}
    \label{eq:VoI_F_from_C}
    VoI_\text{F}(\vb) = \Delta \scC(\vb) + \gamma \sum_{y=1}^{n_Y} \Eps_Y\big(\vt^{\pi_A}(\vb),y\big)~ VoI_\text{F}\big(\Tau_Y^{\pi_A}(\vb,y)\big)
\end{equation}
where the equivalent immediate cost $\Delta \scC$ is:
\begin{equation}
    \label{eq:Delta_C}
    \Delta \scC(\vb)=V^{\pi_A,\pi_B}_W(\vb) - \scU^{\pi_B}_W(\vb) = VoI_\text{C}^\text{O}(\vb) + \Delta V^{\pi_A,\pi_B}_W(\vb)
\end{equation}
function $\Delta V^{\pi_A,\pi_B}_W$ is $(V^{\pi_A,\pi_B}_W - V^{\pi_B}_W)$ and function $V^{\pi_A,\pi_B}_W$ is:
\begin{equation}
    \label{eq:V_pi_A_pi_B}
    V^{\pi_A,\pi_B}_W(\vb) = \scC^{\pi_A}(\vb) + \gamma\sum_{w=1}^{n_W} \Eps_W\big(\vt^{\pi_A}(\vb),w\big)~ V_W^{\pi_B}\big(\Tau_W^{\pi_A}(\vb,w)\big)
\end{equation}
and it represents the value following policy $\pi_A$ at current step, and policy $\pi_B$ from next step on, when additional information is available at all future steps.
Hence, function $\Delta \scC$ represents the impact of two connected contributions: adopting policy $\pi_B$ instead of $\pi_A$ at current step, and collecting and processing additional information, also at current step.
In other words, the value of flow of information can be intended as the expected discounted sum of the optimistic value of current information and of the cost change due to following policy $\pi_B$ instead of policy $\pi_A$, where the system evolves under policy $\pi_A$ and with only background information.

In the special case when the two policies are identical, i.e. $\pi_A=\pi_B$, function $\Delta V^{\pi_A,\pi_B}_W$ is zero, and immediate cost $\Delta \scC$ is equal to $VoI^\text{O}_\text{C}$.
Hence, in that case, $VoI_\text{F}$ accumulates $VoI_\text{C}^\text{O}$.

\section{Optimal value functions and $\alpha-$vectors}
~\label{OptVal}.
While previous formulation is general, we show here how the formulation becomes in the case of optimal values and policies.
To find the optimal value $V^*_Y$ (when additional information is not available), we rewrite Bellman equation Eq.~\ref{eq:V_pi_Y} as:
\begin{equation}
\label{eq:Opt_V_Y}
    V^*_Y(\vb) = \min_{a\in\Omega_A} \Big[  \mathcal{C}(\vb,a) + \gamma\sum_{y=1}^{n_Y} \mathcal{E}_Y\Big(\vt\big(\vb,a\big),y\Big)~ V_Y^*\Big(\mathcal{T}_Y\big(\vb,a,y\big)\Big)\Big]
\end{equation}
where expected immediate cost is $\mathcal{C}(\vb,a) = \sum_{i=1}^{n_X} C(i,a)b(i)$, so that $\mathcal{C}^\pi(\vb)=\mathcal{C}\big(\vb,\pi(\vb)\big)$.
Optimal policy $\pi_Y^*$ is obtained using operator $\text{argmin}$ instead of $\text{min}$ in Eq.~\ref{eq:Opt_V_Y}.

Similarly, we can define value function $V^*_W$ and policy $\pi^*_W$ (with available additional information), and also optimal pre-posterior functions $\scU^*_Y$ and $\scU^*_Z$, following Eq.~\ref{eq:PrePostValue}.

Optimal functions can be expressed, approximately, as the lower envelope of a set of linear functions, each defined by an $n_X$-dimensional column vector, called ``$\alpha$-vector".
Using set $\Gamma_Y = \{\alpha_1, \alpha_2, \dots, \alpha_{H_Y}\}$ of $H_Y$ vectors, then we can express the value function as:
\begin{align}
\label{eq:alpha_vecs}
    V^*_Y(\vb) = \min_{j} ~\alpha_j^\top \vb
\end{align}
Each $\alpha$-vector is also associated with a current action, so that policy $\pi$ can be derived by the dominant vector, as a function of current belief $\vb$, and selecting the corresponding action.

Eq.~\ref{eq:alpha_vecs} shows that set $\Gamma_Y$, together is actions associated with vectors, is sufficient to compute the optimal value function and the corresponding policy.
\cite{PineauSurvey} reviews numerical methods for identifying the set of $\alpha$-vectors.
A basic point-based value iteration method for computing the set of vectors in outlined in \ref{App_PBVI}.

Similarly, we can express function $V^*_W$, accounting for additional information, using another set of $H_W$ $\alpha$-vectors, $\Gamma_W = \{\alpha_1, \alpha_2, \dots, \alpha_{H_W}\}$.

\subsection{VoI under optimal policy }~\label{VoItOpt}
In the optimal setting, value functions are always continuous and concave, because they are the lower envelope of a set of linear functions. Hence, the VoI is always non-negative and continuous.
We can easily show that the value of current information, under optimal behaviour, is non-negative.
Because of total probability rule, the expected posterior belief is equal to the prior one, i.e. $\E_{Z}[\mathbf{u}_Z(\vb,z)]=\vb$, hence, re-writing Eq.~\eqref{eq:VoI_C}, we can express the pessimistic value of current information, $VoI_\text{C}^{*\text{P}}$, as:
\begin{equation}
\label{eq:opt_VoI_C_P}
    VoI_\text{C}^{*\text{P}}(\vb) = V_Y^*\big(\E_Z[\vu_Z(\vb,z)]\big) - \E_Z\big[ V_Y^*\big(\vu_Z(\vb,z)\big)\big] \geq 0
\end{equation}
which is non-negative because of Jensen' inequality, $V_Y^*$ being a concave function, as shown above.
Similarly, the optimistic value of current information, $VoI_\text{C}^{*\text{O}}$ is obtained as in Eq.~\eqref{eq:opt_VoI_C_P}, using instead $V_W^*$ of $V_Y^*$.
Since $V_W^*$ is also concave, $VoI_\text{C}^{*\text{O}}$ is also non-negative.

To prove that the value of flow of information under optimal behaviour, $VoI^*_\text{F}$, is also non-negative, we refer to Eqs.\eqref{eq:VoI_F_from_C},\eqref{eq:Delta_C},\eqref{eq:V_pi_A_pi_B}.
Because of optimally, under persistent availability of additional information we note that $V_W^*$, the optimal value function following optimal policy $\pi_W^*$, is not higher than value $V_W^{\pi_Y^*,\pi_W^*}$ (defined in Eq.\eqref{eq:V_pi_A_pi_B}), obtained adopting policy $\pi_Y^*$ at current step.
Hence, from Eq.\eqref{eq:Delta_C}, we conclude that $\Delta \scC$ is always non-negative. As Eq.\eqref{eq:VoI_F_from_C} shows that $VoI^*_\text{F}$ accumulates immediate non-negative cost $\Delta \scC$, we conclude that $VoI^*_\text{F}$ is also non-negative.

\subsection{Relation among value functions}~\label{VoICons}
\begin{figure}
    \centering
        \includegraphics[scale=0.6]{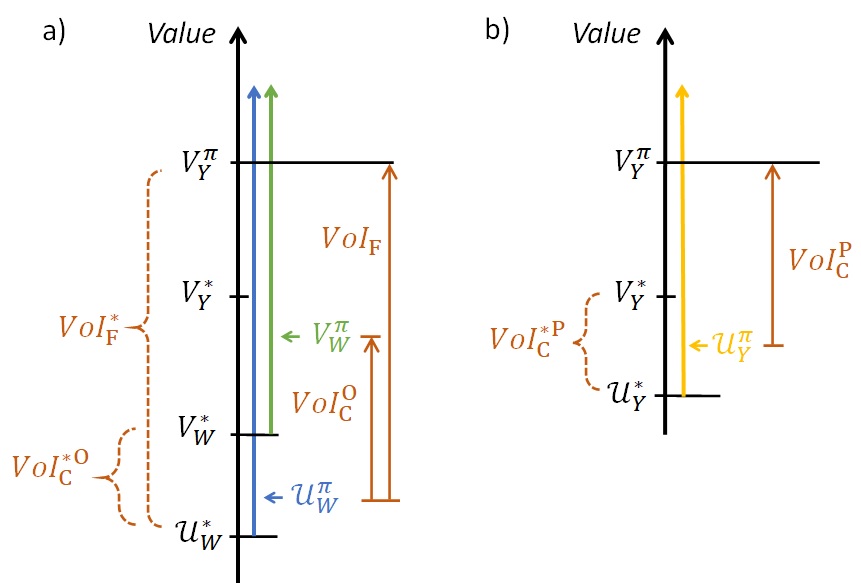}
    \caption{Inequalities among value functions: optimistic value of current information and flow of information (a) and pessimistic value of current information (b).}
    \label{figBounds}
\end{figure}

We outline inequalities among values when an agent adopts the optimal or any other policy.
It is easy observe that 
$$
\forall\vb\in\Omega_B, \forall\pi :  V^\pi_Y(\vb) \geq V^*_Y(\vb) \geq V^*_W(\vb) \geq \scU^*_W(\vb)  $$
Moreover:
$$
\forall\vb\in\Omega_B,  V^*_Y(\vb) \geq \scU^*_Y(\vb) \geq \scU^*_W(\vb)  $$
Fig.~\ref{figBounds} shows these inequalities and the corresponding VoI.

Two value functions that we cannot compare are $\scU^*_Y$ and $V^*_W$ because, depending on the context, the value of having additional information at only present step can be higher or lower of having additional information always from the next step.

\section{Modeling societal regulations}
\label{ModSocReg}
In previous sections we have illustrated how to assess the VoI of additional information, when a policy is assigned.
Now we discuss our assumption for identifying the policy that society assigns to the agent.
To do so, we introduce two different costs functions: $C(x,a)$ models agent's cost, $L(x,a)$ models costs as quantified by ``society", i.e. by the policy makers issuing regulations.
The policy maker identifies the optimal policy related to societal costs, following the approach outlined in Section~\ref{OptVal}.
As such policies depends on the available information, there are two such policies:  $\pi_Y$ is the policy without additional information, and $\pi_W$ that without additional information always available in the future.

Now, the policy maker assigns to the agent these policies, so that the agent is forced to follow them.
We consider two settings.
In the ``fixed" policy approach, the agent must follow the same policy $\pi_Y$, regardless of the availability of additional information.
This is a simpler setting, from standpoint of regulations (as only a single policy needs to be defined).
In the ``flexible" policy approach, instead, the agent follows policy $\pi_Y$ without additional information, but policy $\pi_Z$ if additional information is available in the future.
Usually, policy $\pi_Z$ is less restrictive than $\pi_Y$, and the optimal policies of the agent are even less restrictive (when the agent is more risk taking).
Hence, allowing for a flexible policy, although being a more complicated setting, helps the agent reducing the penalty due to societal restrictions, and also helps society optimizing its goal.

In general, these policies are not optimal for the agent (while they are for the policy maker).
The VoI is assessed by the agent (i.e., using cost function $C$), following the framework outlined in previous Sections.
Instead, when VoI is assessed from societal standpoint, it is non negative, as the selected policy is optimal.

\section{Information avoidance and overvaluation}
\label{IAOE}
If the VoI is negative, the agent finds convenient to avoid information, even if its acquisition is free of costs.
As mentioned in the introduction, we refer to this case as IA.
Instead, if the VoI for the agent following the societal policy is less than the VoI assigned by society itself and less than the VoI assigned by the unconstrained agent, we refer to that case as IOV.
\subsection{Avoidance of current information}
\begin{figure}
    \centering  \includegraphics[scale=0.45]{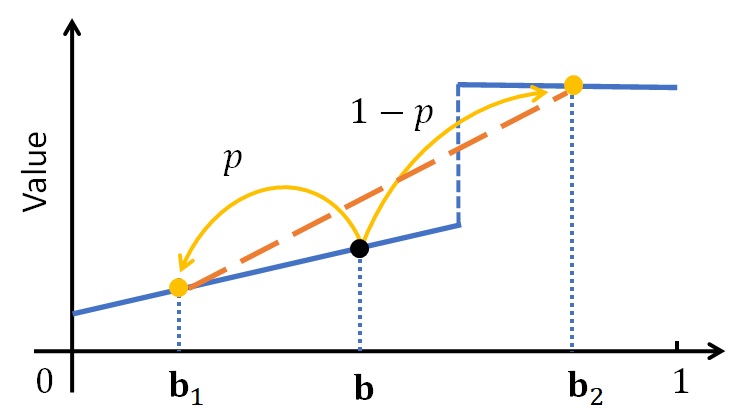}
    \caption{Example of the occurrence of IA with binary additional information.}
    \label{IA_Value}
\end{figure}

As illustrated in Section ~\ref{PolEval}, the value function under non optimal policies can be non-concave and discontinuous.
In this section, we show that for each non-concave value function, there always exists at least one belief and at least one emission function (modeling information) makes the VoI negative, hence triggering IA.
We start with the analysis of current information.
We consider a non-concave value function $\tilde{V}:\Omega_B\rightarrow\R$, so that $ \exists \vb, \vb_1, \vb_2 \in\Omega_B, \exists p \in [0, 1], $
with $\vb = \vb_1 p + \vb_2 (1-p)$
and 
\begin{equation}
    \label{eq:nonConcave}
    \tilde{V}(\vb) < p~\tilde{V}(\vb_1) + (1-p)~\tilde{V}(\vb_2)
\end{equation}
This can be interpreted in relation with the outcome of a binary monitoring system, where $z=1$ can represent an ``silence", and $z=2$ an ``alarm".
Current belief state is $\vb$, and the two possible posterior beliefs are $\vb_1=\vu_Z(\vb,1)$, obtained with probability $p=\Eps_Z(\vb,1)$ and $\vb_2=\vu_Z(\vb,2)$, obtained with probability $\Eps_Z(\vb,2)=1-p$, as illustrated in Fig.~\ref{IA_Value}.
The value of current information (either pessimistic or optimistic, depending on $\tilde{V}$ being $V_Y^\pi$ or $V_W^\pi$, respectively) is therefore negative, because of Eq.\eqref{eq:nonConcave}.

Fig.~\ref{IA_Value} provides an intuition about contexts where IA occurs. If current belief $\vb$ is related to the value at the base of a ``jump", i.e. a discontinuity induced by an epistemic constraint, then the agent may tend to avoid a ``noisy" information source, that can take the posterior belief on the opposite side (i.e. on ``top) of the jump.
\subsection{Overvaluation of current information}
Conversely, strong IOV occurs in a similar setting, when the current belief is already related to the value on top of the jump. In that case, the information, even is highly noisy, offers the opportunity of a significant cost reduction (if the posterior belief falls at the base of the jump), even if the belief updating is barely significant.
This is consistent with what described by \cite{pozzi2020information}, about one-shot decision making problems (there, the loss function plays for role of the value function).
\subsection{Avoidance of flow of information}
To assess the value of flow of information, we start considering the fixed policy setting.
Eqs.\eqref{eq:VoI_F_from_C},\eqref{eq:Delta_C} illustrate how the value of flow of information can be intended as the (discounted, expected) sum of value of current information.
Hence, if the latter is negative for all possible belief, so will be the former (similarly, is the latter is positive, so will be the former).
If, instead, if the latter is positive or negative depending on the belief, then the sign of the former depends on the properties of the stochastic evolution of the process, as indicated by Eq.\eqref{eq:VoI_F_from_C}.
One relevant feature to summarize the process is the expected value of flow of information, where expectation is related to an uncertain current belief.
This quantity depends on the assumed distribution of belief.
Let $p_\infty$ be the asymptotic distribution of the process describing the evolution of the belief, following policy $\pi_A$ and with only background information available.
A numerical procedure for identifying such a distribution in POMDPs is outlined in \cite{li2021predicting}. 
Assuming the process being ergodic, distribution $p_\infty$ is unique.
Hence the expectation of $VoI_\text{F}$ can be expressed as:
\begin{equation}
\label{eq:Exp_VoI}
    \E \big[VoI_\text{F}(\vb)\big] = \frac{1}{1-\gamma}\Big[ \E \big[VoI_\text{C}^\text{O}(\vb)\big] + \E \big[\Delta V^{\pi_A,\pi_B}_W(\vb)\big] \Big]
\end{equation}
This quantity can be either positive or negative, depending on the problem.
For example, in the fixed policy setting, if the expected optimistic value of current information is negative, so is the expected value of flow of information.
If the Markov process is rapidly mixing and the discount factor is high (i.e. with mixing time is much lower than $(1-\gamma)^{-1}$), then we can expect function $VoI_\text{F}$ to be almost insensitive respect to current belief $\vb$, and be close to the expected value computed in Eq.~\eqref{eq:Exp_VoI}.

Generally, we have observed in some analyses that the value of current information tends to be highly sensitive respect to the current belief, while the value of flow of information tend to be less sensitive.
\section{Numerical investigations}
~\label{Examples}
\subsection{A basic example}
\label{BPS}
\begin{figure}
    \centering  \includegraphics[scale=0.1]{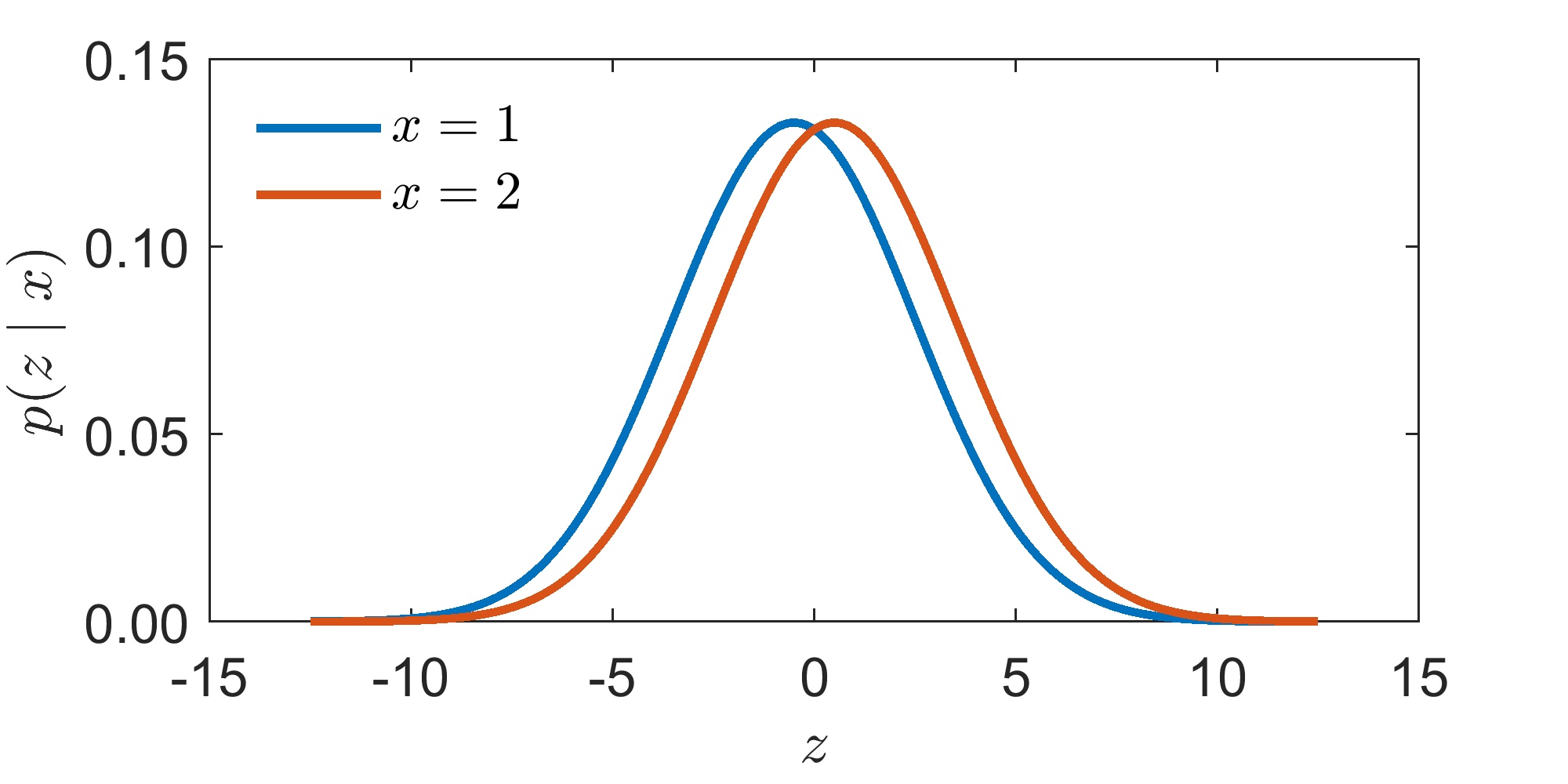}
    \caption{Gaussian emissions for the basic example with $\sigma=3$.}
    \label{figBasicGuassin}
\end{figure}

In this section, we outline a basic example to investigate when and how IA and IOV occur.
Suppose a component has $n_X=3$ physical states: $x=1$ indicates an intact component, $x=2$ a damaged one and $x=3$ a failed one.
Two actions are available, i.e. $n_A=2$, with $a=1$ indicating doing-nothing and $a=2$ repairing the component.
The transition probabilities are reported in Table~\ref{tabEx}, where matrix $\mathbf{T}_a$ reports, in entry $\{i,j\}$, probability $T(i,a,j)$, as defined above.
Those matrices depend on $3$ parameters, modeling the deterioration rates, and for now we set $p_{12} = 0.04$ and $p_{23} = 3~p_{12} = 0.12$.
Background information is such that the failure state is immediately detectable, so observation $Y$ is binary, i.e. $n_Y=2$, and $E_Y(1,a,1)=E_Y(2,a,1)=E_Y(3,a,2)=1$, for any action $a$, while the other entries in table $E_Y$ are zero.
Hence, at every step the belief is a $3$-dimensional vector, but the third entry, after current observation $y$ is processed, is either zero or one. In the former case, the belief can be unequivocally defined by damage probability $P_\text{DAM}$, that is the probability of the current state being $x=2$.
Additional information $z$ is emitted as a normal random variables, with standard deviation $\sigma$, and mean $\mu$ which is $-1/2$ if state is $x=1$, and $1/2$ if $x=2$.
The real axis is discretized in $n_Z=300$ outcomes, to get a finite dimension table $E_Z$. Fig.~\ref{figBasicGuassin} shows the pair of emission functions when $\sigma=3$ (before discretization).

The failure cost for the agent is unitary, so when current action is $a=1$, cost $C$ is $0$ if $x$ is $1$ or $2$, and it is $1$ if $x$ is $3$.
Hence, all other costs can be intended as normalized respect to the failure cost.
If $a=2$, then cost is increased by repair cost $C_R$, here set to $0.5$.
Cost matrix $L$, for society, is identical to agent's cost $C$, except for the repair cost, which is $L_R=0.25$.
The discount factor $\gamma$ is $0.95$.
Hence, by solving the POMDP, we discover that the societal optimal policy $\pi_Y$, without additional information, is to repair if and only if $P_\text{DAM}$ is above threshold value $\chi=35\%$ (or if failure has been detected).
In the fixed policy setting, the agent follows this policy regardless of the availability of additional information.
Note that, as will be illustrated below, the agent would find optimal to repair less often respect to the societal prescription, as her repair cost is higher than the societal one.

\begin{figure}
    \centering  \includegraphics[scale=0.07]{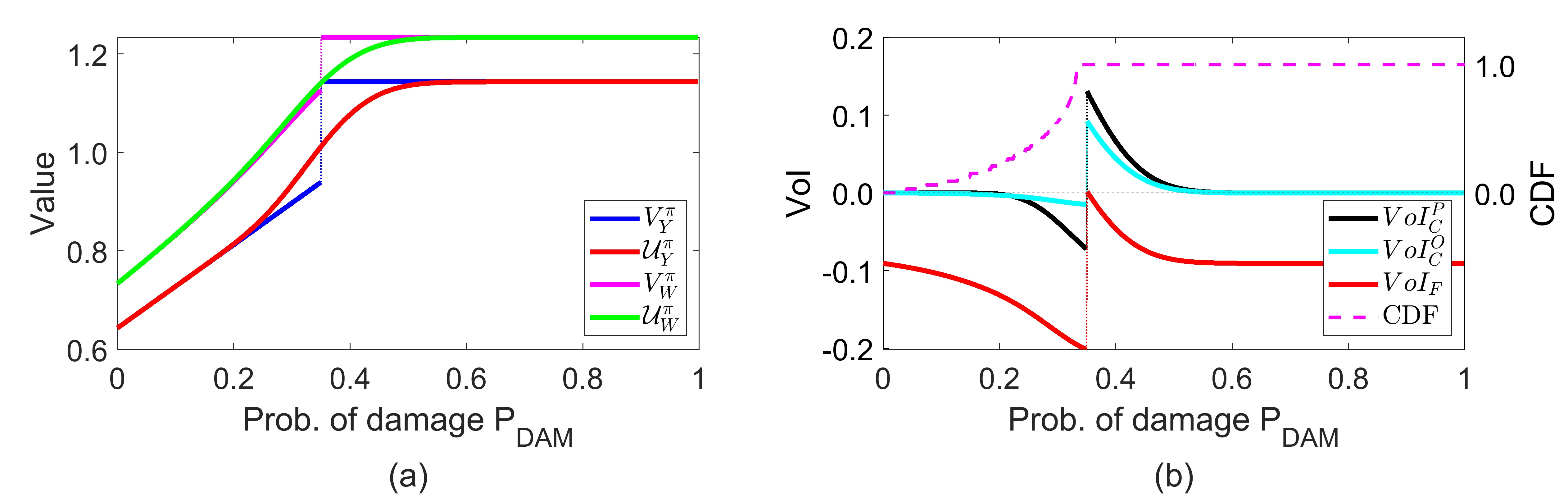}
    \caption{Value (a) and VoI (b) for the basic example.}
    \label{figVandVoI}
\end{figure}

Fig.~\ref{figVandVoI}(a) reports the value functions for the agent, and Fig.~\ref{figVandVoI}(b) the corresponding VoI.
The jump of $V_Y^\pi$ and of $V_W^\pi$ at $\chi=35\%$ occurs because the agent is forced by the social constraint to repair above that threshold.
Instead, pre-posterior value functions are continuous, because so is the random variable defining the additional information $z$.
These latter functions can be seen of smoothed versions of the original value functions (in analogy with a ``moving average smoothing").
The value of current information, either optimistic or pessimistic, is negative for $P_\text{DAM}$ less than threshold $\chi$,
and IA is strong just below it.
At that belief, the pessimistic value of current information is about $0.08$ of the failure cost.
In this setting, in absolute value the optimistic value of current information is always smaller than the pessimistic one and their signs are always consistent.

Fig.~\ref{figVandVoI}(b) also shows the CDF related to the asymptotic probability $p_\infty$, showing that the belief process (when only background information is available), tends to stay under threshold $\chi$.
Because of this, the expected value of $VoI_\text{C}^\text{O}$ is negative, and it is equal to $-0.008$.
The expected value of $VoI_\text{F}$ is $-0.16$, using \eqref{eq:Exp_VoI}.

\begin{figure}
    \centering  \includegraphics[scale=0.07]{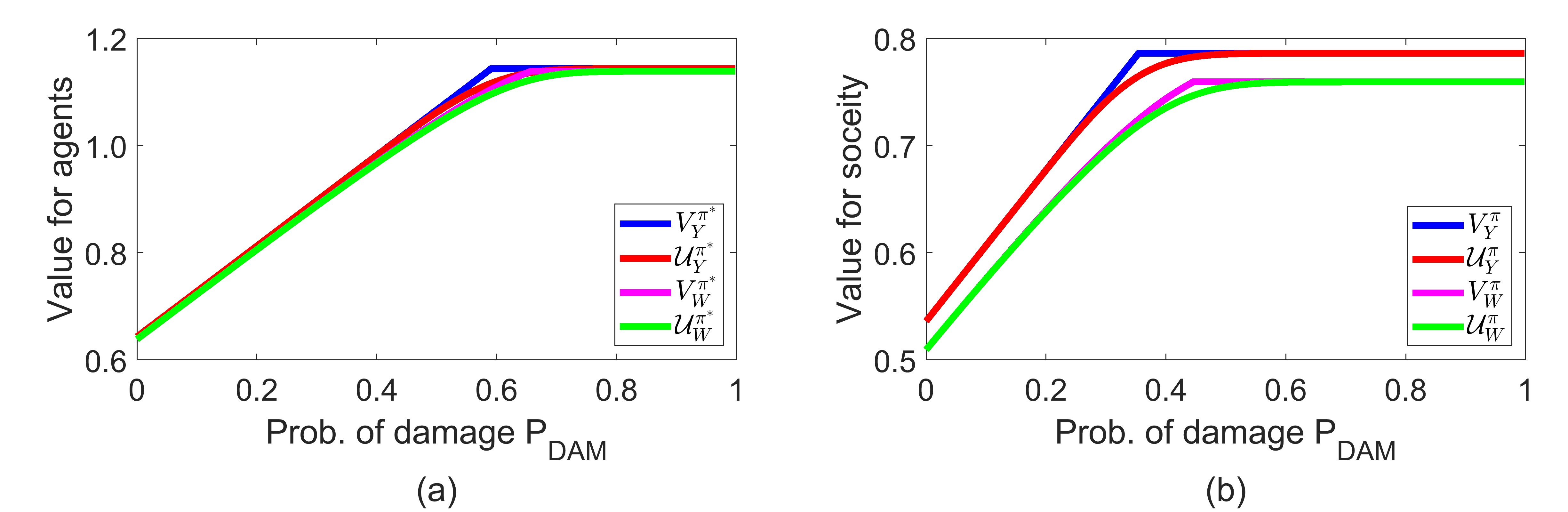}
    \caption{Optimal value for agent (a) and for society (b).}
    \label{figVAS}
\end{figure}

\begin{figure}
    \centering  
\includegraphics[scale=0.07]{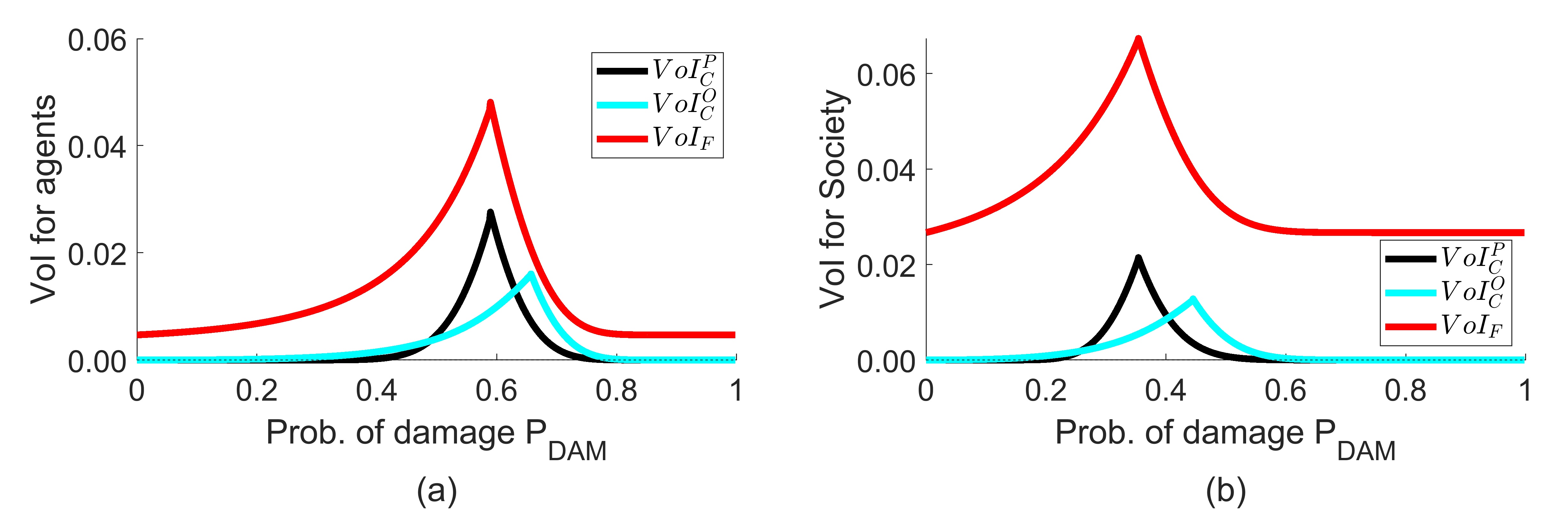}
    \caption{VoI under optimal policies for agent (a) and for society (b).}
    \label{figVoIVS}
\end{figure}

As a comparison, Fig.~\ref{figVAS} shows the optimal value function for agent (a) and for society (b) when they can follow their optimal corresponding policies.
Threshold $\chi$ for repairing is higher for the agent acting unconstrained (it is about $60\%$).
The corresponding value of information is shown in Fig.~\ref{figVoIVS}.
Comparing Fig.~\ref{figVoIVS} and Fig.~\ref{figVandVoI}(b), we note that the VoI under epistemic constraint is well above the unconstrained one, for damage probability just above $\chi$: this is a case of IOV.

\begin{table}
\footnotesize
\centering
\caption{ Transition and observation matrix for numerical examples\label{tabEx}}
\begin{tabular}{ c  c c}
\toprule
$ ~~~~\mathbf{T}_{1} =
\begin{bmatrix}
1-p_{12} & p_{12} & 0\\
0 & p_{23} & 1-p_{23}\\
0 & 0 & 1
 \end{bmatrix} $ 
 $ ~~~~\mathbf{T}_{2} =
\begin{bmatrix}
1-p_{12} & p_{12} & 0\\
1-p_{12} & p_{12} & 0\\
1-p_{12} & p_{12} & 0
 \end{bmatrix} $ \\
\bottomrule
\end{tabular}
\end{table}

\begin{figure}
\centering
 \includegraphics[scale=0.07]{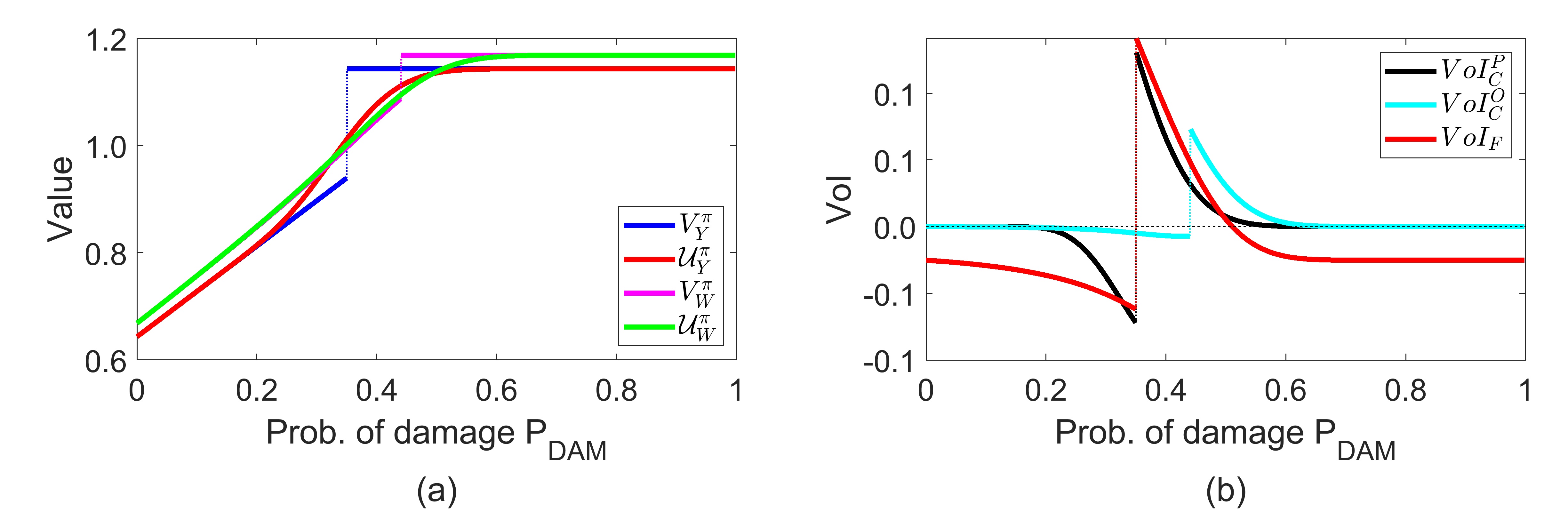}
    \caption{Value (a) and VoI (b) following flexible policies $\pi_A$ and $\pi_B$.}
    \label{figflex1}
\end{figure}

\begin{figure}
    \centering
        \includegraphics[scale=0.07]{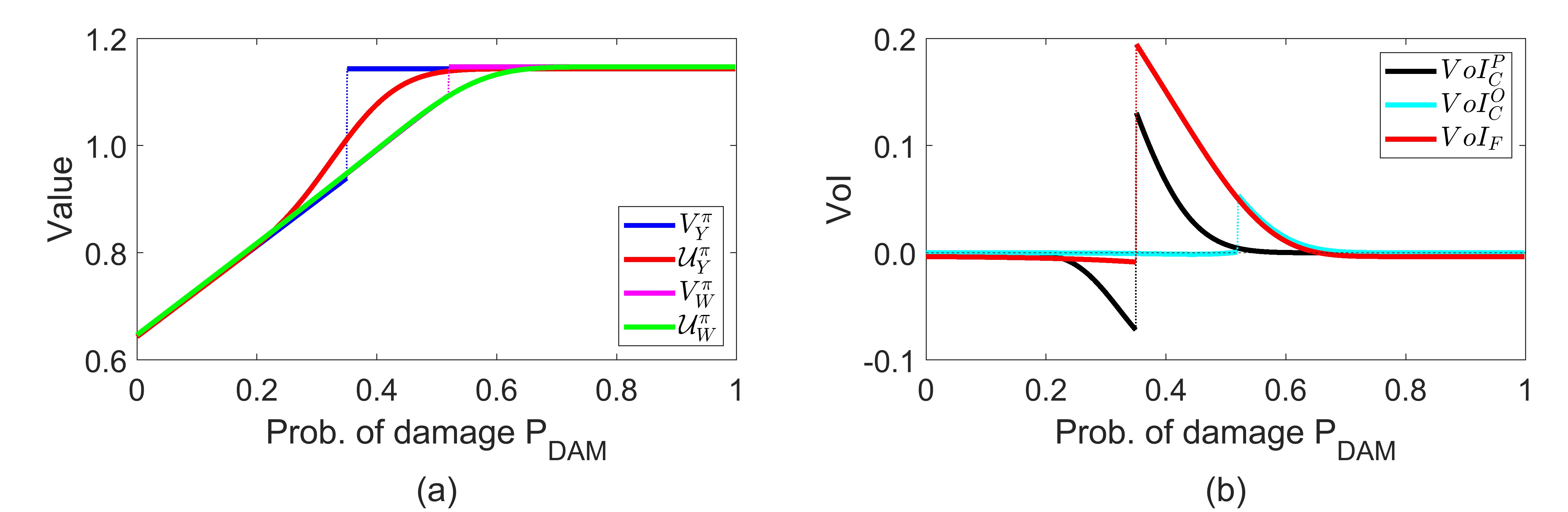}
    \caption{Value (a) and VoI (b) following flexible policies $\pi_A$ and $\pi_C$.}
    \label{figflex2}
\end{figure}

\subsection{Comparing fixed and flexible settings}
While previous example referred to a fixed policy imposed by society to the agent, we now investigate the effect of allowing for a flexible policy.
In this new setting, we compute also the optimal policy for society when additional information is always available.
We now define $\pi_A$ as the optimal societal policy without additional information, with repair threshold at $\chi=35\%$, as seen before.
$\pi_B$ is the optimal societal policy with additional information when $\sigma= 3$, and the resulting repair threshold is $\chi=44\%$.
$\pi_C$ is the corresponding policy when $\sigma= 1$, and the repair threshold becomes $\chi=52\%$.

Following the flexible setting, an agent repair less often (than in the fixed policy case) when additional information is available, as the threshold is higher.
Fig.~\ref{figflex1} and ~\ref{figflex2} show the value function (a) and VoI (b) when $\sigma= 3$ and when $\sigma= 1$ (so that the policy with additional information is $\pi_B$ and $\pi_C$, respectively).
We see that IA is stronger in the fixed policy case (shown in Fig.~\ref{figVandVoI}), because the flexible policy allows the agent to behave more closely to her optimal policy (as noted before, the optimal threshold is $\chi=60\%$ for the unconstrained agent). 

\subsection{Parametric analysis}
In this section, we investigate how the VoI is affected by the parameters of the problem. 

\subsubsection{Noise level and VoI}
\begin{figure}
    \centering
         \includegraphics[scale=0.08]{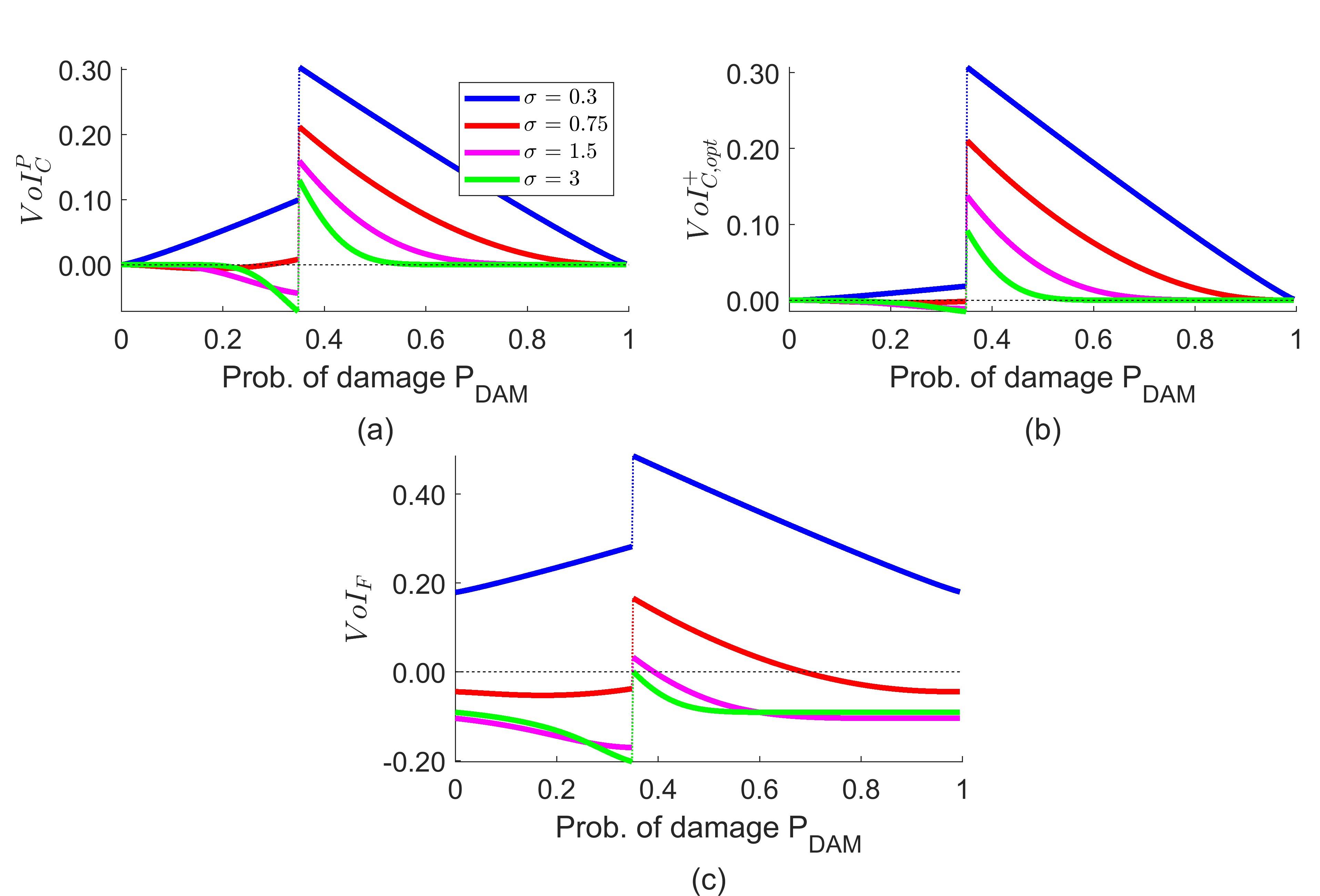}
    \caption{Pessimistic value of current information (a),  Optimistic value (b), value of flow of information (c), depending on noise level.}
    \label{figGuassPlots}
\end{figure}

\begin{figure}
    \centering
 \includegraphics[scale=0.05]{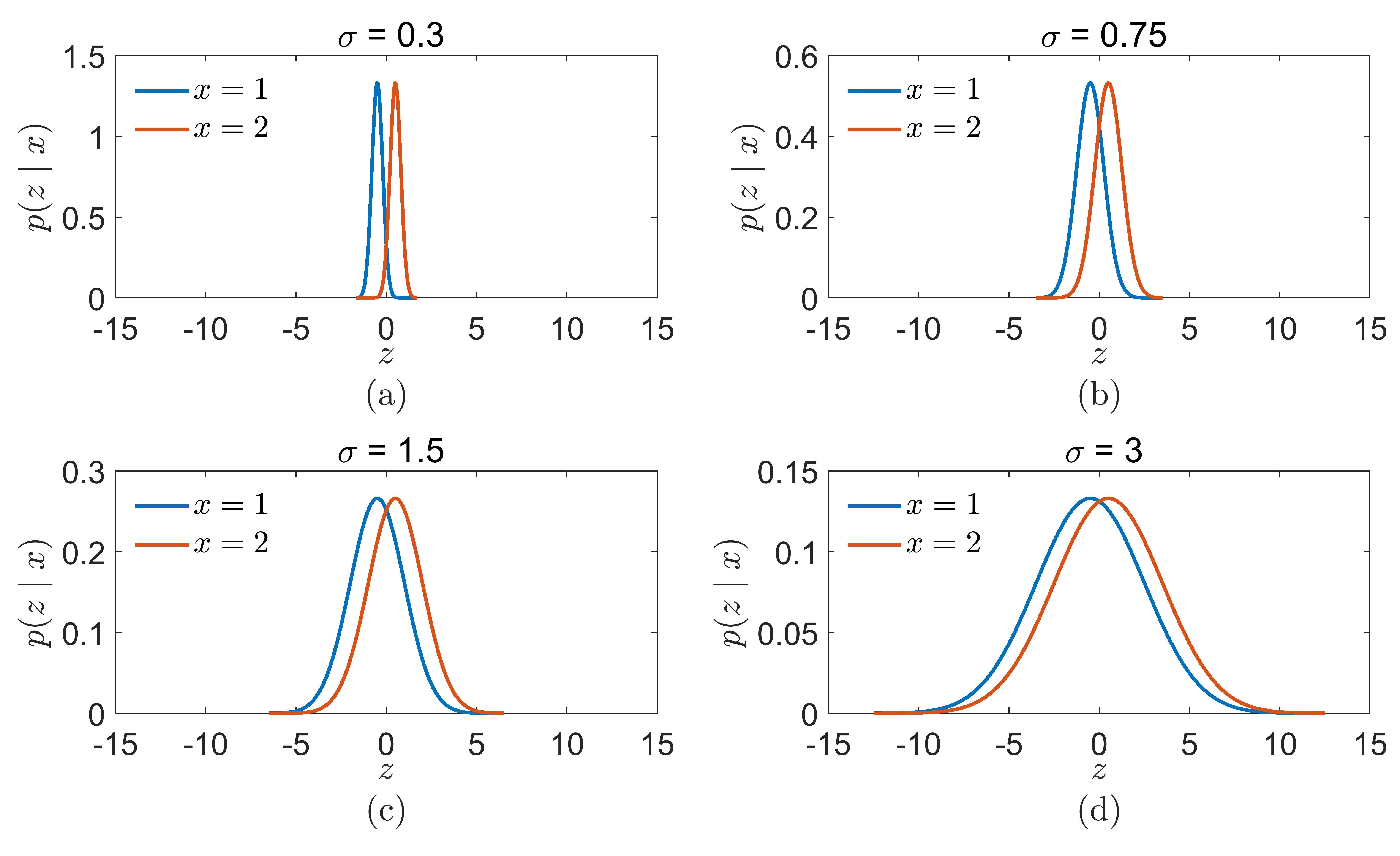}        
    \caption{Gaussian emissions for $\sigma=0.3$ (a), $\sigma=0.75$ (b), $\sigma=1.5$ (c) and $\sigma=3$ (d).}
    \label{figGuassPlots2}
\end{figure}

\begin{figure}
    \centering
         \includegraphics[scale=0.2]{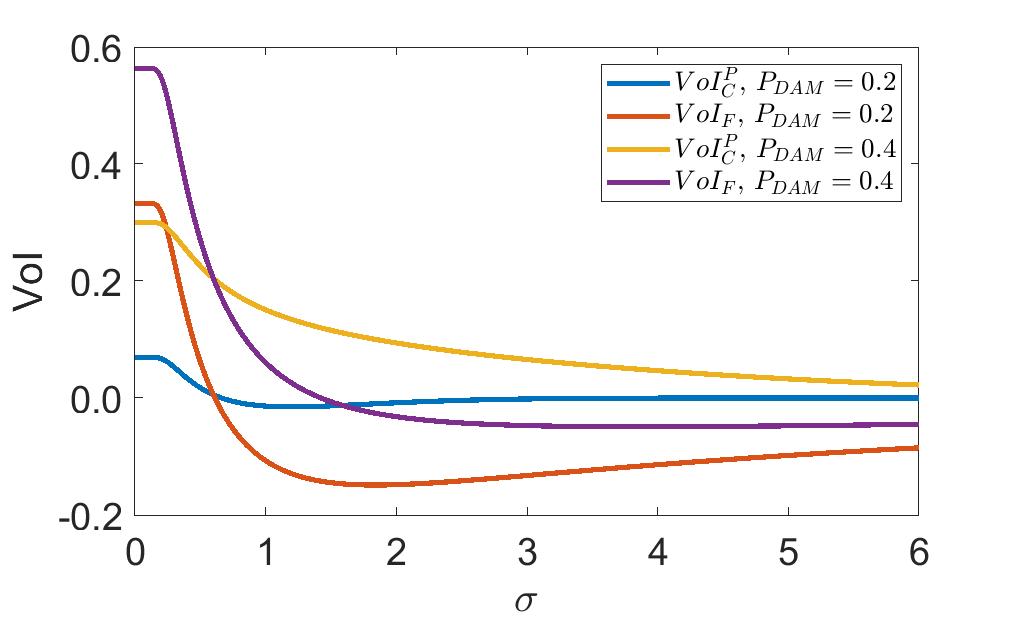}
    \caption{VoI \textit{vs} noise level.}
    \label{figEps}
\end{figure}

Fig.~\ref{figGuassPlots} illustrates how the VoI changes depending on noise level $\sigma$ for the additional information $z$.
Four values of $\sigma$ are considered, and the corresponding emission probabilities are illustrated in Fig.~\ref{figGuassPlots2}.
We plot the relation between VoI and $\sigma$ in Fig.~\ref{figEps}, for two beliefs: at $P_\text{DAM}=20\%$ and at $P_\text{DAM}=40\%$ (with zero failure probability).
Mostly, more precise information has higher value, but there are exceptions, and the relation is not always monotonic. For $\sigma$ going to infinity, we expect the VoI going to zero, except when the belief is close to the jumps induced by the thresholds.

\subsubsection{Deterioration rate and VoI}
\begin{figure}
    \centering
 \includegraphics[scale=0.092]{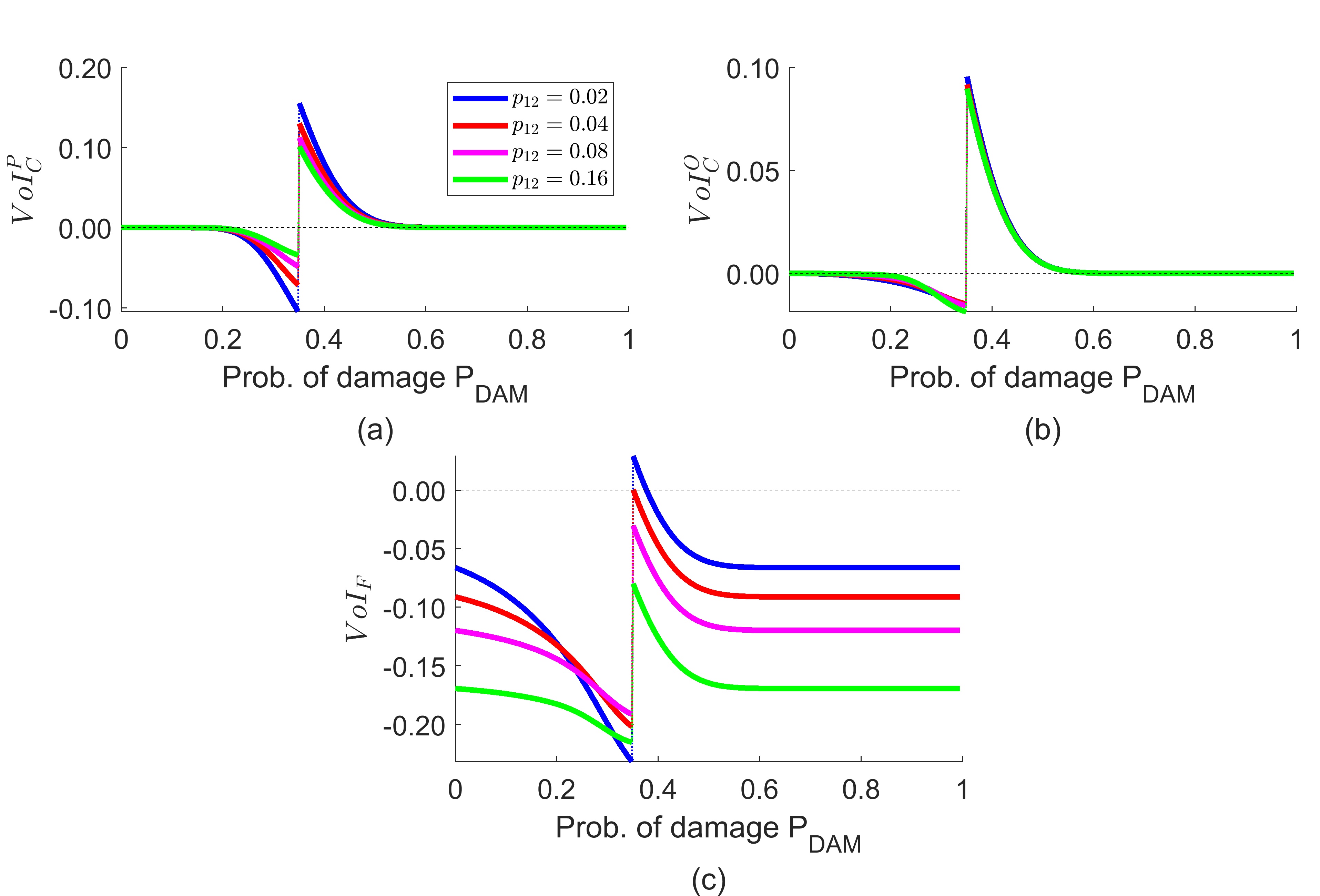}        
     \caption{Pessimistic value of current information (a),  Optimistic value (b), value of flow of information (c), depending on degradation rate.}
    \label{figRates}
\end{figure}

Fig.~\ref{figRates} illustrates the relation between the VoI and the deterioration rate, modeled by transition probability $p_{12}$, keeping $p_{23}=3~p_{12}$ and the repair threshold at $\chi=0.35$.
We investigate the transition probability $p_{12}$ at value $2\%$, $4\%$, $8\%$ and $16\%$.
$VoI^\text{P}_\text{C}$ is monotonically decreasing (in absolute value) with degradation rate, while $VoI^\text{O}_\text{C}$ is almost insensitive to the degradation rate.
The change of $VoI_\text{F}$ with degradation rate is more complicated and not monotonic for $P_{\text{DAM}}$ less than $\chi$.

Generally, the IA is stronger when the degradation rate is higher.
The change in $VoI_\text{F}$ is due to the (small) change in $VoI^\text{O}_\text{C}$, but also to the change in the asymptotic distribution $p_\infty$: when $p_12$ grows, it becomes more probable that the belief is close to right side of $\chi$, where $VoI^\text{O}_\text{C}$ is highly negative.

\subsubsection{Repair threshold and VoI}
\begin{figure}
    \centering
 \includegraphics[scale=0.12]{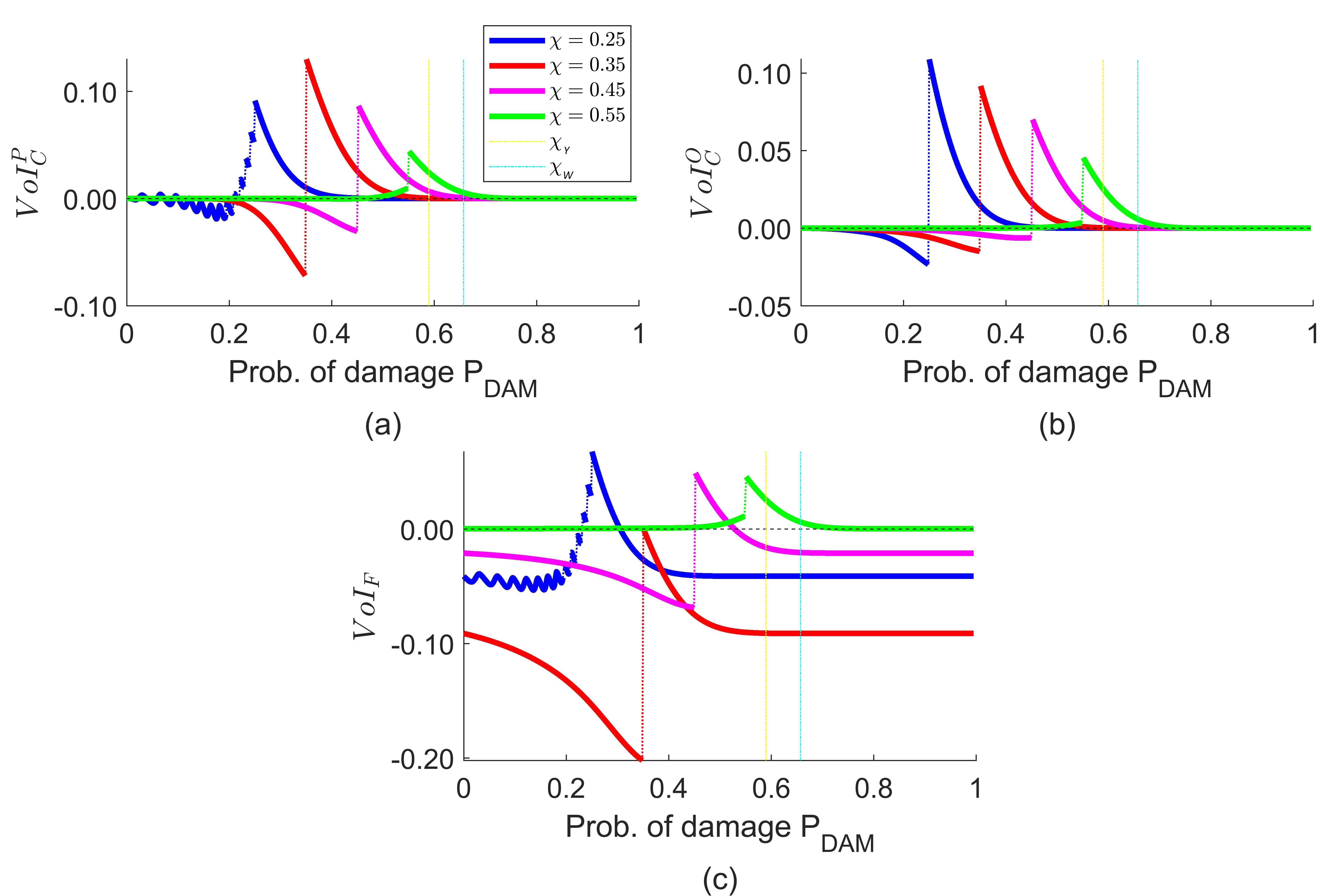}
    \caption{Pessimistic value of current information (a),  Optimistic value (b), value of flow of information (c), depending on repair threshold.}
    \label{figConst}
\end{figure}

The last parametric analysis is shown in Fig.~\ref{figConst}.
By changing the societal cost for repairing, $L_R$, we change the imposed repair threshold $\chi$.
The graph illustrates how $\chi$ influences the agent VoI (the vertical lines report optimal threshold for the unconstrained agent without, $\chi_Y$, and with, $\chi_W$, additional information).
For the optimistic value of current information, the lower $\chi$ the higher IA and IOV.
However, for the pessimistic value of current information of the value of flow of information, the most intense IA is at $\chi=35\%$, the value investigated in the basic setting.

% ...

\subsection{Details on numerical implementation}
\label{DetNumImpl}
We provide some details of the numerical implementation.
In the problem introduced in Section \ref{BPS}, belief domain is $\Omega_B=\{\vb: \exists P_\text{DAM}\in [0,1]: \vb=[1-P_\text{DAM},P_\text{DAM},0]^\top\}\cup[0,0,1]^\top$, and it is made up by a segments where $b(3)=0$ and by the failure points.
A set of $n_B=1001$ representative beliefs is distributed in $\Omega_B$, evenly spaced in the log-scale of $P_\text{DAM}$.
The point-based algorithm described in \ref{App_PBVI} is iterated $n_T=180$ times, so that $\gamma^{n_T}<10^{-4}.$
The repair cost for society, $L_R$, is set to $0.246$, so that the optimal repair threshold is $\chi=0.35$.
In that algorithm, a new identified $\alpha$-vector is added to the set only if the minimum euclidean distance respect to all vectors already in the set is higher than $3\varepsilon_V$.
When only background information is available, only $n_H=2$ $\alpha$-vectors are in set $\Gamma_Y$, regardless of $\varepsilon$.
This is because the optimal societal policy is to do nothing until the background information is ``silence", that is until no failure is detected (and to repair if failure is detected). Following a sequence of silences, damage probability $P_\text{DAM}$ converges to $1/3$, that is lower than the repair threshold.
When additional information is also available, the cardinality of set $\Gamma_W$ depends on $\varepsilon_V$: it is $n_H=18$ if $\varepsilon_V=0.01$, $n_H=129$ if $\varepsilon_V=10^{-3}$ and $n_H=507$ if $\varepsilon_V=10^{-4}$.
The numbers of vectors are similar when also the additional information is available.
For this problem, the outcomes are almost indistinguishable when using the Grid Method outlined in \cite{li2021predicting} based, again, on $n_B=1001$ representative beliefs.

\section{Conclusions}
~\label{Concl}
In this article, we have investigated the occurrence of IA and IOV in sequential decision making under epistemic constraints as those imposed by societal constraints.
The results extend those about one-shot decisions, presented in \cite{pozzi2020information}.
Respect to that case, the sequential decision making case is computationally more challenging and it is related to the evaluation of non optimal policies, that can be done using approaches based on FSCs.
Also, the definition of VoI is richer, as it includes the value of current and of flow of information.
By illustrating how these values are related, we have shown that IA and IOV can occur both for current and for flow of information.
The value function plays, in the sequential decision making, the role played by the loss function in one-shot decision making, and the occurrence of IA and IOV is related to the properties of these functions, specifically to non-concavity and discontinuity.

The topic of mitigating IA and IOV, designing and adopting mechanisms of incentives and penalty has been investigated, for a simple setting of one-step decision making, in \cite{pozzi2020information}. 

\appendix

\section{Proof of Eq.\eqref{eq:VoI_F_from_C}}
\label{App_Proof}
To prove Eq.\eqref{eq:VoI_F_from_C}, we can express function $V_Y^{\pi_A}$, starting from Eq.\eqref{eq:V_pi_Y}, adding and subtracting the same term and using the definition of $V^{\pi_A,\pi_B}_W$, from Eq.\eqref{eq:V_pi_A_pi_B}, to get:
%
\begin{equation}
\label{eq:V_pi_Y_proof}
\begin{split}
    V_Y^{\pi_A}(\vb) = \scC^{\pi_A}(\vb) + \gamma\sum_{y=1}^{n_Y} \Eps_Y\big(\vt^{\pi_A}(\vb,y)\big)~ V_Y^{\pi_A}\big(\Tau_Y^{\pi_A}(\vb,y)\big) + \\ \gamma\sum_{w=1}^{n_W} \Eps_W\big(\vt^{\pi_A}(\vb,w)\big)~ V_W^{\pi_B}\big(\Tau_W^{\pi_A}(\vb,w)\big) - \\ \gamma\sum_{w=1}^{n_W} \Eps_W\big(\vt^{\pi_A}(\vb,w)\big)~ V_W^{\pi_B}\big(\Tau_W^{\pi_A}(\vb,w)\big) \\
    = V^{\pi_A,\pi_B}_W(\vb) +  \gamma\sum_{y=1}^{n_Y} \Eps_Y\big(\vt^{\pi_A}(\vb,y)\big)~ V_Y^{\pi_A}\big(\Tau_Y^{\pi_A}(\vb,y)\big) - \\ \gamma\sum_{w=1}^{n_W} \Eps_W\big(\vt^{\pi_A}(\vb,w)\big)~ V_W^{\pi_B}\big(\Tau_W^{\pi_A}(\vb,w)\big)
\end{split}
\end{equation}
To develop the last sum in previous equation, we note that we can express operator $\Tau_W$ in terms of operators $\Tau_Y$ and $\vu_Z$:
\begin{equation}
    \Tau_W\big(\vb,a,w\big) = \vu_Z\big(\Tau_Y(\vb,a,y),z\big)
\end{equation}
and emission operator $\Eps_W$ in terms of $\Eps_Y$ and $\Eps_Z$:
\begin{align}
    \Eps_W(\vb,w) = \Eps_Y(\vb,y) \Eps_Z(\vb,z)
\end{align}
Using the last two equations and Eq. \eqref{eq:PrePostValue}, we get:
\begin{equation}
\begin{split}
    \label{eq:V_pi_Y_bis}
    \sum_{w=1}^{n_W} \Eps_W\big(\vt(\vb,a),w\big)~ V_W^\pi\big(\Tau_W(\vb,a,w)\big) = \\
    \sum_{y=1}^{n_Y} \Eps_Y\big(\vt(\vb,a),y\big)\sum_{z=1}^{n_Z} \Eps_Z\big(\vt(\vb,a),z\big) ~ V_W^\pi\big(\vu_Z(\Tau_Y(\vb,a,y),z)\big) = \\
    \sum_{y=1}^{n_Y} \Eps_Y\big(\vt(\vb,a),y\big) ~\scU^\pi_Z\big(\Tau_Y(\vb,a,y)\big)
\end{split}
\end{equation}
Substituting the last identity in Eq.\eqref{eq:V_pi_Y_proof}, and grouping common terms, we get: 
\begin{equation}
\begin{split}
    V_Y^{\pi_A}(\vb) = V^{\pi_A,\pi_B}_W(\vb) + \\ \gamma\sum_{y=1}^{n_Y} \Eps_Y\big(\vt^{\pi_A}(\vb,y)\big)~ \Big[V_Y^{\pi_A}\big(\Tau_Y^{\pi_A}(\vb,y)\big) - \scU^{\pi_B}_Z\big(\Tau_Y^{\pi_A}(\vb,y)\big)\Big]
\end{split}
\end{equation}
Finally, by subtracting term $\scU^{\pi_B}_Z(\vb)$ on both sides, we get Eq.\eqref{eq:VoI_F_from_C}.

\section{Point-based Value Iteration}
\label{App_PBVI}
As noted in Sec.\ref{OptVal}, for a POMDP the optimal value function can be represented as the lower envelope of a set of linear functions of the belief, each defined by an $\alpha$-vector.

In this Section we provide a basic procedure to identify a set of $\alpha$-vectors, given the parameters of the POMDP, and then the corresponding policy graph and joint graph.
We keep a subscript indicating that the results refers to what information is available, and $Y$ refers to the availability of background information only.
The procedure belongs to the family of point-based value iteration methods \cite{AIMA}, \cite{PineauSurvey}.
We start assuming that we do have a set of $M$ $\alpha$-vectors, $\Gamma_Y=[\valpha_1,\valpha_2,\dots,\valpha_M]$, representing, via the lower envelope of the corresponding linear functions, the optimal value function at the next step.

Our goal is to ``update" the set into a set of $M'$ $\alpha$-vectors, $\Gamma'_Y=[\valpha_1,\valpha_2,\dots,\valpha_M']$, describing the value function at current step.
If we are able to update set $\Gamma_Y$ into set $\Gamma'_Y$, we can iterate this updating step until the set becomes stationary, according to some criteria of tolerance.

To formulate the updating step, we refer to map 
$m_Y(\vb):\Omega_B\rightarrow \{1,2,\dots,M\}$, that identifies the index of the dominating $\alpha$-vector in set $\Gamma_Y$, for any belief $\vb$, and it is defined as:
as $m_Y(\vb)=\text{argmin}_i [\valpha_i^\top\vb]$, as defined in Sec.\ref{NumApprPolEv}.

Note that the value from next step, at a function of next belief $\vb'$, is $\text{min}_i[\valpha_i^\top\vb']=\valpha_{g(\vb')}^\top\vb'$.
The ``quality" function, as a function of current belief $\vb$ and current action $a$ is:
\begin{equation}
    Q(\vb,a)=\scC(\vb,a)+\gamma\sum_{y=1}^{n_Y}\Eps_Y\big(\vt(\vb,a),a,y\big)\valpha_{m_Y(\Tau_Y(\vb,a,y))}^\top\Tau_Y(\vb,a,y)
\end{equation}
this function defines the expected discounted accumulated cost when action $a$ is taken, and then the optimal policy takes control from next step.
The stationary optimal policy can be defined, as function of this function, as: 
$\pi_Y^*(\vb)=\text{argmin}_a Q(\vb,a)$.

We now assume to have a set of $n_B$ representative beliefs $\tilde{\Omega}_B=\{\vb_1,\vb_2,\dots,\vb_{n_B}\}$.
As we can identify the optimal action for each representative belief (by minimizing the quality function $Q$ respect to the action), we can re-write symbol  $\pi_Y^*$ to indicate the optimal policy as a function of the inner state:
$\pi_Y^*(h)=\pi_Y^*(\vb_h)$.

The corresponding updating function is obtained as:
\begin{equation}
    \eta^\pi_Y(h,y)=m_Y\Big(\Tau_Y\big(\vb_h,y,\pi(h)\big)\Big)
\end{equation}
If $h$ is the current inner state, $a_h=\pi(h)$ is the current action and $\eta^\pi_Y(h,y)$ identifies the next $\alpha$-vector depending on next information $y$, then the value function $v_h$ depends on current belief $\vb$ as:
\begin{equation}
    v_h(\vb) = \scC(\vb,a_h) + \gamma \sum_{y=1}^{n_Y} \Eps_Y\big(\vt(\vb,a_h),a_h,y\big)~\valpha_{\eta^\pi_Y(h,y)}^\top \Tau_Y(\vb,y,a_h)
\end{equation}
so that current optimal value function at representative belief $\vb_h$, $V_Y(\vb_h)$, is $v_h(\vb_h)$.
Using Eq.\eqref{eq:u_Z}, previous formula reduces to:
\begin{equation}
    v_h(\vb) = \vc_{a_h}^\top\vb + \gamma \sum_{y=1}^{n_Y} \valpha_{\eta^\pi_Y(h,y)}^\top \text{Diag}[\ve_Y(y,a_h)] \vt(\vb,a_h)
\end{equation}
where $\vc_a$ is the column vector of immediate costs for action $a$, so that entry $i$ of the vector is $C(i,a)$.
Expressing the transition operator using the transition matrix, we get:
\begin{equation}
    v_h(\vb) = \vc_{a_h}^\top\vb + \gamma \sum_{y=1}^{n_Y} \valpha_{\eta^\pi_Y(h,y)}^\top\text{Diag}[\ve_Y(y,a_h)]\mT_{a_h}^\top\vb
\end{equation}
We know recognize that function $v_h$ is linear in the current belief, and be expressed as: $v_h(\vb)=\valpha_h'^{\top}\vb$, where:
\begin{equation}
\label{eq:upAlphaVec}
    \valpha_h' = \vc_{a_h} + \gamma \mT_{a_h} \sum_{y=1}^{n_Y} \text{Diag}[\ve_Y(y,a_h)]\valpha_{\eta^\pi_Y(h,y)}
\end{equation}
Eq.\eqref{eq:upAlphaVec} allows us to generate a new $\alpha$-vector for representing the current optimal value function, based on optimality at representative belief $\vb_h$.
By analyzing all representative beliefs, we populate new set  $\Gamma_Y'=[\valpha'_1,\valpha'_2,\dots,\valpha'_{M'}]$.

The updating process outlined here is an application of the value iteration approach (where value at one step is computed from that at the next one).
As, for the discounted infinite horizon process, the value function is time invariant, the approximation of the value iteration algorithm should converge to the actual function.
Hence, after a number of iterations that depends on $\gamma$, we expect that the set of $\alpha$-vectors converges to a stationary set, representing the actual optimal value function.
By construction, the maximum number of vectors is $n_B$, but it may be the case that fewer vectors are sufficient.
To prune vectors, we can check if any pair of identified vectors are identical (possibly because policy and updating functions are identical, for the corresponding pair of representative beliefs), or if they are approximately identical, and delete the redundant vector, as mentioned in Sec.~\ref{DetNumImpl}.
As reviewed also in \cite{li2021predicting}, each inner state $h$ represents a ``conditional plan", and function $v_h$ the corresponding value function following this plan.

Also, each inner state is a node in the policy graph \cite{li2021predicting}, showing how conditional plans are defined recursively.
The ``joint graph" is derived from the policy graph and the ``transition graph", and the corresponding transition matrix is obtained as in Eq.\eqref{eq:T_tilde_pi_Y}.
All $\alpha$-vectors can be derived as value of nodes, solving Eq.\eqref{eq:PolEval}: the value at node $w=\{x,h\}$ is $\alpha_h(x)$.

Our key task for the VoI analysis under constraints is to evaluate a sub-optimal policy. We identify policy $\pi_Y$, which is optimal respect to societal cost function.
Then, we compute the transition matrix $\tilde{\mathbf{T}}^\pi_Y$, in the joint graph.
Now, for any arbitrary immediate agent's costs (even if $\pi_Y$ is not optimal for these costs), we compute cost vector $\tilde{\vc}^\pi$ and, solving linear system Eq.\eqref{eq:PolEval}, we compute nodal values $\tilde{\vv}^\pi_Y$, and then the (sub-optimal) value $V^\pi_Y$ function by Eq.\eqref{eq:pieceWiseLinValue}.

Previous analysis can be repeated when additional information is always available in the future, obtaining policy $\pi_W$, updating function $\eta^\pi_W$ set of $\alpha$-vectors $\Gamma_W$, joint graph transition matrix $\tilde{\mathbf{T}}^\pi_W$, nodal values $\tilde{\vv}^\pi_W$ and value function $\tilde{\vv}^\pi_W$, using emission function $E_W$ instead of $E_Y$.

\section*{Acknowledgement}
The authors thank the support of NSF project SES \#1919453, titled “Attitude towards Information in Multi-agent settings: Understanding and Mitigating Avoidance and Over-Evaluation”.

{
\bibliography{refs}
}

\end{document}